%% file: acl_latex.tex
\pdfoutput=1

\documentclass[11pt]{article}

\usepackage[final]{acl}

\usepackage{times}
\usepackage{latexsym}

\usepackage{booktabs}
\usepackage{multirow}

\usepackage[T1]{fontenc}

\usepackage[utf8]{inputenc}

\usepackage{microtype}

\usepackage{inconsolata}

\usepackage{graphicx}
\usepackage{amsmath}
\usepackage{mathrsfs}
\usepackage{enumitem}

\usepackage{pifont}
\newcommand{\circnum}[1]{\ding{\numexpr171+#1}}

\usepackage{tabularx}   
\usepackage{multirow}   
\usepackage{array}      
\usepackage{booktabs}   
\usepackage{enumitem}   

\usepackage{tcolorbox}
\usepackage{caption}
\usepackage{float}

\tcbset{colback=white, colframe=black, fonttitle=\bfseries}

\title{Domain-Specific Data Generation Framework for RAG Adaptation }

\author{
Chris Xing Tian\textsuperscript{1}\thanks{Equal contribution.},
Weihao Xie\textsuperscript{2}\footnotemark[1],
Zhen Chen\textsuperscript{2},
Hui Liu\textsuperscript{2},\\
\textbf{Zhengyuan Yi\textsuperscript{2},
Haoliang Li\textsuperscript{2},
Shiqi Wang\textsuperscript{2},
Siwei Ma\textsuperscript{3}}\\[6pt]
\textsuperscript{1}Peng Cheng Laboratory, Shenzhen, China\quad
\textsuperscript{2}City University of Hong Kong, Hong Kong SAR\\
\textsuperscript{3}Peking University, Beijing, China\\[6pt]
\texttt{txsing@live.com, swma@pku.edu.cn}\\
\texttt{\{weihaxie-c, zchen979-c, liuhui3-c\}@my.cityu.edu.hk}\\
\texttt{\{zhengyyi, haoliang.li, shiqwang\}@cityu.edu.hk}
}

\begin{document}
\maketitle
\begin{abstract}
Retrieval-Augmented Generation (RAG) combines the language understanding and reasoning capabilities of large language models (LLMs) with external retrieval to produce domain-grounded responses. Effectively adapting RAG systems to domain-specific settings requires specialized, context-rich training data beyond general-purpose question-answering datasets. Here, we propose RAGen, a scalable and modular data-centric framework for generating domain-grounded question–answer–context (QAC) triples tailored to diverse RAG adaptation strategies. These QAC triples serve as training signals for multiple RAG adaptation approaches; in this work, we demonstrate their use for contrastive fine-tuning of embedding models and supervised fine-tuning of LLMs under retrieved contexts. RAGen generates QAC triples by identifying key concepts within documents, producing diverse questions guided by Bloom’s Taxonomy–inspired principles, and pairing them with precise answers extracted from relevant contexts. Its modular pipeline incorporates semantic chunking, hierarchical concept extraction, multi-chunk retrieval, and curated distractor contexts to encourage robust reasoning. Designed for scalability, RAGen efficiently handles large and evolving document corpora without redundant processing, making it particularly suitable for dynamic domains like enterprise knowledge bases.
\end{abstract}



\section{Introduction}

With the growing adoption of large language models (LLMs) in enterprise and organizational settings, there is increasing demand for integrating these models into domain-specific workflows \cite{chiarello2024future, qian2024evolution}. However, concerns over data privacy, regulatory compliance, and the high cost of commercial API usage often prevent organizations from deploying proprietary, cloud-hosted LLMs. As a result, many turn to open-source, locally deployed small- and medium-scale LLMs for internal use.

Despite their accessibility, smaller models inherently suffer from limited language understanding and reasoning capabilities compared to frontier LLMs \cite{chen2024benchmarking, mallen2022not}. This performance gap motivates the use of Retrieval-Augmented Generation (RAG) \cite{lewis2020retrieval}, which supplements an LLM with a retriever to provide external, context-specific information. RAG offers a practical and modular solution for grounding LLM outputs in proprietary knowledge bases without requiring massive model sizes.


However, simply applying off-the-shelf RAG pipelines to new domains often leads to suboptimal performance \cite{barnett2024seven}, as general-purpose retrievers and generators are not aligned with domain-specific terminology or data distributions. This makes RAG adaptation essential. We define RAG adaptation as the process of refining individual components of the RAG pipeline—such as the retriever, embedding model, and LLM—to better match the target domain and improve end-to-end performance \cite{siriwardhana2023improving, liu-etal-2025-unraveling}. In practice, such adaptation is typically achieved through additional domain-specific supervision, for example by fine-tuning embedding models with contrastive objectives or LLMs with question–answer data. Crucially, this supervision can be derived from a small, representative, and potentially desensitized subset of source documents used to generate training data, enabling the use of powerful proprietary models for adaptation without requiring full corpus exposure.


Recent work has explored adapting RAG systems by targeting individual components with specialized training strategies. For example, RAFT \cite{RAG-LLM:RAFT} introduces distractor-aware fine-tuning to improve the robustness of LLMs under noisy retrieval, while inference-time methods such as Self-RAG \cite{RAG-InferTime:SelfRAG} and Open-RAG \cite{RAG-InferTime:OpenRAG} focus on teaching LLMs when and how to invoke retrieval during generation. Other efforts similarly concentrate on improving either the retriever or the generator in isolation through tailored objectives and training pipelines. 

While effective within their respective scopes, these approaches are inherently component-centric: each targets a single module of the RAG pipeline and is tightly coupled to a specific training or inference paradigm and typically assume the availability of specific training data which limits their generalizability across domains and architectures.


To address these limitations, we propose RAGen, a scalable and modular framework for generating high-quality, domain-specific training data to support multi-component RAG adaptation. RAGen is explicitly data-centric: rather than introducing new model architectures or training objectives, it automatically synthesizes domain-grounded Question–Answer–Context (QAC) triples by identifying document-level concepts, assembling multi-chunk evidence, and generating questions guided by Bloom’s Taxonomy-inspired principles \cite{krathwohl2002revision}. These QACs can serve as generic supervision for adapting multiple RAG components, such as contrastive fine-tuning of embedding models and supervised, context-aware fine-tuning of LLMs. Owing to its modular design and reliance on concept-centered evidence rather than fixed schemas, RAGen scales naturally to large, evolving corpora and is well-suited to practical deployment settings such as enterprise knowledge bases and scientific domains.



Empirical results across multiple domains demonstrate that RAGen-generated data significantly improve both retrieval quality and generation accuracy. Compared to baselines, our approach yields deeper, more holistic questions and enhances performance across a variety of adaptation tasks. These findings highlight RAGen as a practical and generalizable solution for building robust, domain-adapted RAG systems.

\begin{figure*}[t]
  \centering
  \includegraphics[width=\textwidth]{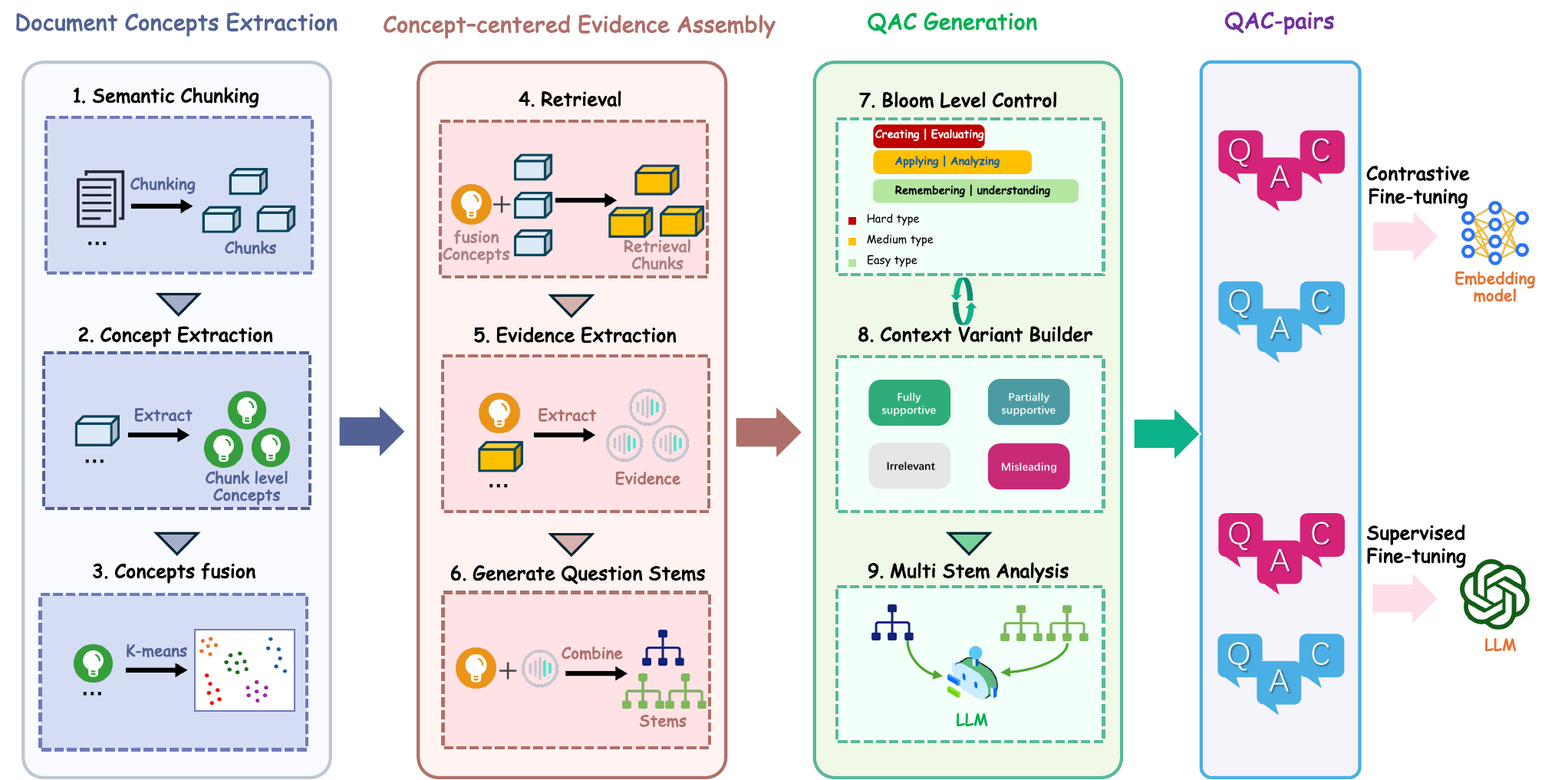}
  \caption{Overview of RAGen framework, a three-stage process that first extract document concepts and then construct question stems, and finally create Question-Answer-Context datasets.}
  \label{fig:overall-flow}
\end{figure*}

\section{Related Work}

\paragraph{Question Generation}
Automatic QA generation has been widely studied to reduce annotation cost and support domain-specific modeling. CliniQG4QA \cite{QAG:CliniQG4QA} generates controlled clinical QA pairs via template-guided phrase prediction and answer-aware generation, but assumes short, clean passages and does not scale to long, noisy enterprise documents. E2EQR \cite{hwang2024explainable} constructs multi-hop questions by iteratively rewriting simpler queries, yet lacks explicit mechanisms for evidence selection and grounding, which are critical for retrieval-augmented settings. FinTextQA \cite{QAG:FintextQA} targets financial text using semantic retrieval and sentence windowing, but depends on external question banks, limiting applicability in unseen or low-resource domains. QAG \cite{ushio2023empirical} adopts an answer-first pipeline to generate multi-hop questions, but is not designed to align question construction with retriever training or RAG-specific supervision.

Several recent question generation frameworks focus on evaluating RAG systems. RAGEval \cite{QAG-Eval:RagEval} introduces the DRAGONBall dataset and associated metrics via a schema–configuration–document–QRA pipeline, producing scenario-driven synthetic corpora for controlled benchmarking. RAGAS \cite{es-etal-2024-ragas} provides reference-free evaluation metrics and a synthetic test set generator based on knowledge-graph construction and evolutionary question rewriting. While effective for probing diverse query behaviors, these methods are primarily designed for evaluation and do not expose persistent semantic concepts or structured context roles.

In contrast, RAGen is explicitly data-centric and targets \emph{RAG adaptation rather than evaluation}. It operates directly on raw, schema-free corpora and generates structured Question–Answer–Context (QAC) units grounded in document-level concepts and multi-chunk evidence which later serve as generic supervision for adapting multiple RAG components.

\paragraph{Retrieval-Augmented Generation (RAG)}
Retrieval-Augmented Generation (RAG) \cite{lewis2020retrieval} enhances language models by grounding generation in externally retrieved documents. A standard RAG pipeline comprises three components: a retriever that selects relevant passages, an embedding model that maps queries and documents into a shared space, and a language model that synthesizes answers from retrieved content.

Prior work has explored improving individual components of this pipeline. Dense retrieval methods such as DPR \cite{RAG-Retriever:DPR} and embedding adaptation approaches like MAFIN \cite{RAG-Embedding:Mafin} focus on enhancing retrieval quality. Other methods, including GraphRAG \cite{RAG-Embedding:GraphRAG}, model inter-passage relationships using structured representations, but often rely on predefined schemas that limit flexibility across domains. 


On the generation side, RAFT \cite{RAG-LLM:RAFT} introduces distractor-aware supervision to improve the model’s robustness against noisy or irrelevant contexts. More recent work has focused on inference-time retrieval control, where the LLM actively guides what and when to retrieve. Representative approaches include Self-RAG \cite{RAG-InferTime:SelfRAG}, OpenRAG \cite{RAG-InferTime:OpenRAG}, and R1Searcher \cite{RAG-InferTime:R1Searcher} which adopt end-to-end training paradigms to align retrieval behavior with generative intent.

Despite their effectiveness, these methods are largely \emph{component-centric} and assume the availability of high-quality, domain-specific training data tailored to particular objectives or modules. In contrast, our work addresses this upstream data bottleneck. We propose \textbf{RAGen}, a data-centric framework that automatically generates semantically grounded Question–Answer–Context (QAC) datasets from raw corpora. RAGen provides reusable supervision that can be flexibly applied to train and adapt multiple RAG components, enabling end-to-end improvement across diverse architectures and domains without assuming task-specific annotations.

\section{Methodology}

The RAGen pipeline is designed to automatically generate rich, high-quality question–answer–context (QAC) training data to support diverse RAG adaptation strategies. RAG adaptation refers to the process of systematically refining individual components of a Retrieval-Augmented Generation (RAG) system—such as the large language model (LLM), retriever, and embedding model—to enhance accuracy and robustness of the RAG system under dynamic domain-specific settings.

In the following, we will present the RAGen workflow, which comprises three main modules: \textit{(i) Document concepts extraction}, \textit{(ii) Question stems construction}, and \textit{(iii) QA and context generation}. The overall workflow is illustrated in Fig.\ref{fig:overall-flow}.

\subsection{Document Concepts Extraction}
\paragraph{Semantic chunking.}
Given the domain documents $D$, we employ the standard \textit{llamaindex chunker} to partition the text into a set of coherent chunks $\{d_1,d_2,\dots\}$.

\paragraph{Chunk–level concept extraction.}
For each chunk $d_i$, ChatGPT-4o \cite{openai-4o} is prompted to extract a set of concise, non-generic descriptors referred to as chunk-level concepts: 
$\mathcal{C}_{i}=\{\,c_{1}^{i},c_{2}^{i},\dots\}$, which capture the central themes of $d_i$.

\paragraph{Concept Fusion.}
To capture high-level semantics across a document, all chunk-level concepts are further fused based on semantic similarity, resulting in a de-duplicated set of representative \emph{document-level concepts}:
$O = \{o_1, o_2, \dots, o_K\}.$

The fusion process begins by eliminating redundant terms and synonyms from the chunk-level concepts. Each remaining concept is then embedded into a vector space using the OpenAI Ada embedding model \cite{openai-ada}. Finally, the K-means clustering algorithm is applied to group these embeddings into K semantically coherent clusters, where K serves as a tunable hyperparameter. For each cluster, the concept closest to the centroid is selected as its representative, serving as a concept at the document level. Alternatively, an LLM-based summarization can be employed to abstract each cluster into a concise descriptor as the document-level concept.

This fusion step reduces the chunk-level concept space to a compact set of document-level themes, which guide cross-chunk retrieval and enable holistic, globally grounded question generation. While related in spirit to hierarchical retrieval methods such as RAPTOR \cite{sarthi2024raptor}, our approach performs a single, non-recursive clustering over LLM-extracted concept phrases to obtain interpretable document-level concepts used solely as semantic anchors for offline QAC generation, rather than for inference-time retrieval.

\subsection{Concept–centered Evidence Assembly}
\paragraph{Cross-chunk Retrieval.}
Given the document-level concepts derived in the previous stage, we perform cross-chunk retrieval to collect semantically relevant contexts. For each concept, we use a retriever–reranker pipeline consisting of the dense retriever and \textit{BGE-Reranker-Base} \cite{zhang2024soaring} to retrieve the top-$N$ most relevant chunks from the document corpus. Due to the abstract and high-level nature of document-level concepts, this process often surfaces non-sequential chunks scattered across the document. This enables a departure from traditional single-chunk-based generation strategies, which tend to produce overly localized contexts and shallow questions. Instead, our approach supports the synthesis of holistic, multi-faceted questions grounded in distributed evidence.

\paragraph{Evidence Extraction.}
Although the retrieved chunks are semantically related, they are often coarse-grained and may contain information unrelated to the target concept. To isolate relevant content, we perform sentence-level filtering within each chunk to extract a concept-focused subset of text, referred to as the evidences $e$, via sentence window retriever, denoted as $ d \xrightarrow[]{o_i} \{ e_0^{o^i}, e_1^{o^i}, \ldots, e_N^{o^i} \} $. This step simulates the human annotation process, where a reader selects specific spans of interest before crafting a question. By narrowing the scope to concept-relevant sentences, we ensure that the subsequent question generation process remains focused, interpretable, and controllable.

Unlike existing QA generation methods that operate on isolated, single chunks, our approach assembles evidences from multiple, non-contiguous chunks scattered across the document. The resulting set of evidences for each concept forms a semantically grounded \emph{Question Stem}, denoted as $\mathcal{S}$, which serves as the basic unit for downstream question generation.

While single-stem inputs enable the generation of concept-focused, context-aware questions, we further support multi-stem combinations—allowing the question generator to condition on multiple concepts simultaneously. This enables the creation of global, cross-concept questions that require deeper reasoning and more complex logical chaining. As such, our approach supports the generation of holistic, semantically rich questions that go beyond the limitations of single-chunk-based methods, better simulating human-level comprehension and reasoning over long-form content.

\subsection{QAC Generation}
\paragraph{Bloom's question-type.}
After constructing a list of $K$ question stems, each consisting of concept-centered evidence, we sample them to form input to the question generator. We define the number of stems combined per input as the combination level, denoted by $\ell$. When $\ell = 1$, we iterate through all individual stems. For $\ell \geq 2$, the number of possible combinations becomes $C_{K}^{\ell}$, which can grow rapidly. To manage this combinatorial explosion, we impose an upper limit on the number of questions generated for each level $\ell$; once this threshold is met, we stop enumerating further combinations at that level.
For each input consisting of one or more Question Stems, we prompt ChatGPT-4o to generate diverse types of questions supported by the associated evidences. To guide this process, we adopt Revised Bloom’s Taxonomy \cite{krathwohl2002revision}, a widely used pedagogical framework that categorizes cognitive learning objectives in ascending order of complexity:
\begin{itemize}[nosep]
    \item \textit{Remembering}: Recognizing or recalling information,
    \item \textit{Understanding}: Constructing meaning from information.
    \item \textit{Applying}: Using knowledge in new situations,
    \item \textit{Analyzing}: Breaking down information into parts and finding evidence,
    \item \textit{Evaluating}: Making judgments based on criteria,
    \item \textit{Creating}: Putting elements together to form a coherent whole.
\end{itemize}

By aligning question types with Bloom’s Taxonomy, we simulate the cognitive learning trajectory of humans and enable the generation of questions that span from factual recall to complex synthesis and reasoning. This approach allows us to explicitly control the difficulty distribution of the generated dataset, ensuring a balanced mix of lower-order and higher-order cognitive questions. In addition, the flexible combination of stems—especially at higher $\ell$ levels—naturally promotes diversity in both content and reasoning depth, enabling the dataset to cover a wider range of topics and inferential patterns.

Notably, for combinations where $\ell \geq 2$, it is possible that no meaningful question can be inferred—particularly when the concepts in the stems are semantically unrelated. In such cases, we discard the current combination and move on to the next.

By combining chunk-level concept fusion with multi-stem aggregation, our framework supports both cross-chunk and cross-concept reasoning. This layered design promotes the generation of high-quality, pedagogically diverse, and cognitively rich question–answer–context samples suitable for domain-specific RAG adaptation.

\paragraph{Question Generation.}
Conditioned on the selected stem combination and Bloom’s Taxonomy levels, we prompt ChatGPT-4o\footnote{https://platform.openai.com/docs/models/gpt-4o-mini} to generate the question, its reference answer, a concise reasoning trace, and the supporting evidences.

To enhance retrieval sensitivity and robustness, we further associate each question–answer (QA) instance with four curated context variants (Below, we use the question "\textit{what are the possible colors of apple?}" as the example):
\begin{itemize}[nosep]
    \item \textit{Fully-supportive}: Sentences directly drawn from the evidence set that completely answer the question. Example: "\textit{Apples have various colors: red, green, yellow depending on the variety.}"
    \item  \textit{Partially-supportive}: A subset of the evidence that contains incomplete information, requiring cross-evidence reasoning. Example: "\textit{Fuji apples are famous for their red surface.}"
    \item  \textit{Irrelevant}: Content from the same domain but unrelated to the question. Example: "\textit{Bananas turn from green to yellow when they ripen.}"
    \item  \textit{Misleading}: Topically related but semantically insufficient content that could plausibly mislead a reader. Inspired by human reading comprehension distractors, these passages share surface similarity but fail to answer the question. Example: "\textit{Apple trees have flowers that are mainly white or light pink.}"
\end{itemize}
Unlike prior methods that rely solely on randomly sampled chunks as distractors, our well-curated distractors increases the semantic difficulty of the retrieval task while encourages higher-order reasoning and a deeper understanding of domain semantics during model adaptation.

Through the RAGen pipeline, we finally generate high-quality, domain-specific datasets from seed documents to support a variety of RAG adaptation strategies. Each data sample includes a question, the associated concepts, a corresponding answer, and multiple curated contexts. These elements collectively enable fine-grained control over question difficulty and content diversity.

\section{Experiments}
We evaluate the proposed RAGen framework by constructing three domain-specific datasets: PPFS, TradePolicy, and AIBusiness. PPFS is derived from APEC Policy Partnership on Food Security meeting documents covering topics such as water management, rural development, and sustainable agriculture. TradePolicy includes import/export regulations (primarily for meat and seafood) collected from eight APEC economies. BusinessAI consists of technical reports on AI adoption across various business sectors. All data are collected from publicly available websites.

We generate QAC datasets from these seed documents using RAGen and compare them against two baselines: 1. AutoRAG\cite{kim2024autorag}: an automated framework that searches for optimal RAG pipeline configurations on user-provided data, including a built-in dataset generation module. 2.	LlamaIndex Dataset Generator\cite{llamaindex_data_gen}: an open-source QA data generator for RAG evaluation. We refer to it as LlamaIndex in this paper.

Both baselines follow a single-chunk question generation paradigm: AutoRAG uses a simplified Bloom-style taxonomy (factual/conceptual), while LlamaIndex applies intra-chunk retrieval similar to our evidence extraction step. We exclude RAGEval due to its reliance on structured schemas, which are incompatible with our unstructured corpora.

Each dataset is constructed from self-contained documents, enabling standalone QA generation without cross-document reasoning. Evaluation splits are shown in Table~\ref{tab:dataset-stats}. We apply the same document partitions and maintain comparable question volumes across RAGen, AutoRAG, and LlamaIndex to ensure fairness.

To assess the impact of RAGen data, we conduct experiments on both embedding model customization and LLMs fine-tuning using 4$\times$NVIDIA RTX 3090 GPUs. Results consistently show that RAGen-generated datasets lead to improved performance across multiple adaptation settings.

\begin{table}
  \centering
  \begin{tabular}{lllll}
    \hline
    \textbf{Domain} & \textbf{Corpus No.} & \textbf{Questions No.}  \\
    \hline
    PPFS &  15 /3 &         2726 /2502 /2084 \\
    TradePolicy      &  20 /5 &         1977 /1820 /1500                     \\
    BusinessAI      &  17 /3 &          2228 /2118 /2072  \\
    \hline
  \end{tabular}
  \caption{Corpus size (training/evaluation) and number of generated questions (RAGen / LlamaIndex / AutoRAG) for each domain.}
  \label{tab:dataset-stats}
\end{table}

\paragraph{Hyperparameter discussion} During question generation, all methods segment documents into 1024-token chunks with a 200-token overlap. For single-chunk baselines (AutoRAG, LlamaIndex), question generation is controlled by a single hyperparameter: the number of questions per chunk. However, this approach is inherently constrained by the limited semantic scope of each chunk, and increasing the value often leads to redundant or low-quality questions. To balance question quantity and quality, we carefully tune this hyperparameter for both baselines. As shown in Table~\ref{tab:dataset-stats}, AutoRAG consistently produces the fewest questions across all domains.

In contrast, RAGen generates questions from document-level concept stems, which reflect higher-level semantics across chunks. The number of stems scales with content richness, and RAGen further supports multi-stem combinations, enabling cross-concept, cross-chunk reasoning. To ensure fairness, we restrict generation to combination levels $\ell \leq 2$ with a cap of 50 questions for $\ell$ = 2. Even under this constraint, RAGen consistently yields more diverse and semantically rich questions than single-chunk methods.

\vspace{-4pt}
\subsection{Dataset Analysis}
\begin{figure}[t]
  \centering
  \includegraphics[width=\columnwidth,clip]{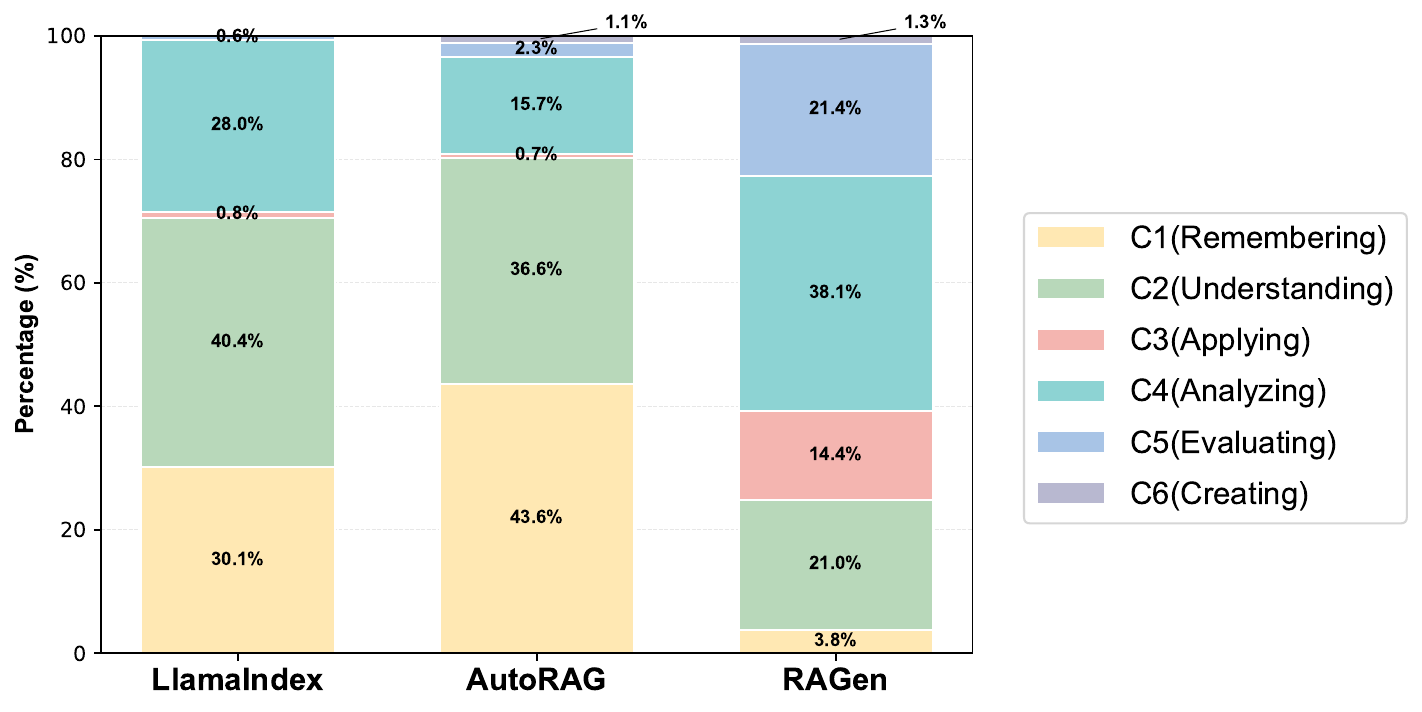}
  \caption{Cognitive level distribution in PPFS domain.}
  \vspace{-6pt}
  \label{fig:bloom}
\end{figure}

\paragraph{Cognitive Level Coverage.} Fig.\ref{fig:bloom} shows the distribution of Bloom’s cognitive levels for questions generated by LlamaIndex, AutoRAG, and RAGen. Compared to the other two, RAGen produces a markedly richer mix of higher-order question types (Analyzing, Evaluating, Creating) while drastically reducing low-level (Remembering and Understanding) questions. This indicates that RAGen-generated data are more holistic and conceptually comprehensive, moving beyond surface-level recall to support deeper reasoning and complex learning objectives—essential for building robust, domain-adapted RAG systems.

\vspace{-6pt}
\paragraph{Cross-concept and Cross-chunk Questions.}
RAGen supports multi-stem conditioning, where multiple document-level concepts—each associated with evidence from distinct chunks—are jointly used to generate a single question. This design naturally enables the generation of cross-concept questions, which often span multiple chunks, resulting in more holistic and semantically rich QA pairs. As illustrated in Fig.~\ref{fig:cross-concept}, such questions require deeper reasoning and capture relationships across disparate parts of a document. In contrast, single-chunk methods like LlamaIndex are limited to localized questions, reducing both answer completeness and dataset diversity. RAGen’s ability to support multi-faceted, cross-concept reasoning reflects a key advantage for developing realistic RAG systems.

\begin{figure}[htbp]
\centering
\begin{tcolorbox}[width=\columnwidth, colback=gray!6, colframe=black,
                  boxsep=4pt, left=4pt, right=4pt, top=4pt, bottom=4pt]
\small
\textbf{Question:} How can the integration of document drafting agents impact the incremental profit and loss in life sciences companies? (Concept: Document Drafting Agent \& Profit and Loss)
\\[3pt]
\textbf{Evidence(from Chunk4):} […agents could free up 25 to 40 percent of employees’ workloads… allowing employees to focus on more strategic, value-adding, and productive work…]
\\[3pt]
\textbf{Evidence(from Chunk6):} […the full potential of enterprise-wide agentic transformation could boost top medtechs’ EBITDA by 2.2 to 4.7 percentage points…]
\\[3pt]
\textbf{Evidence(from Chunk11):} […documentation agents can achieve 75 to 80 percent productivity gains for initial document generation…]
\\[3pt]
\hrulefill
\textbf{Answer:} \circnum{1} …can automate the generation of manufacturing practice documents, achieving productivity gain. \circnum{2} …allow employees to focus on more strategic tasks, potentially freeing up their workload. \circnum{3} …the full potential of enterprise-wide agentic transformation could boost top medtechs’ EBITDA...
\end{tcolorbox}
\caption{Cross-concept question sample: By drawing on 3 non-adjacent evidence sources, cross-concept questions promote deeper, more holistic reasoning, moving beyond localized facts to capture broader operational and financial implications.}
\label{fig:cross-concept}
\end{figure}

\subsection{Embedding model customization}
\label{sec:embedding_adaptation}
In domain-specific RAG systems, the embedding model plays a pivotal role in retrieval accuracy, which directly influences generation quality. While pre-trained models provide general-purpose embeddings, they may underperform in specialized domains. For example, the word “pitch” carries very different meanings in sports domain (“The baseball pitch was perfect.”) and business domain (“The startup delivered a great pitch to investors.”), illustrating how domain context shapes semantic interpretation. To address this, we follow prior works in embedding fine-tuning \cite{wang2023improving, bge_embedding, llm_embedder} and adopt the open-source framework \textit{FlagEmbedding} \cite{flagembedding} for both embedding model fine-tuning and evaluation to investigate how fine-tuning embedding models through contrastive training with synthetic domain-specific data can enhance retrieval performance in domain-adapted RAG settings.

\paragraph{Setup.} We conducted embedding model customization experiments on all three domain-specific datasets. To demonstrate the effectiveness of the RAGen datasets, we select three different embedding models: BGE-large-v1.5 \cite{bge-large-v15} (hereafter referred to as BGE-large), BGE-m3 \cite{chen2024bge}, and E5-large-v2 \cite{wang2022text}. Under the InfoNCE objective \cite{oord2018representation}, we set the learning rate to 1e-5 for 3 epochs, with the temperature parameter  \(\tau = 0.02\) and the number of negative samples set to 2 . All models are fine-tuned using a full-parameter training setup with consistent hyperparameters across all runs. For evaluation, we assess the fine-tuned model on the split-out evaluation datasets of the three domains. Specifically, for each domain, we randomly select 300 samples from the AutoRAG, LlamaIndex, and RAGen evaluation datasets respectively to form the final evaluation set. We adopt Recall@K (K=1, 5, 10) and Mean Reciprocal Rank (MRR@10) as the evaluation metrics, which are widely used in the evaluation of information retrieval system.

For all methods, we construct contrastive training triplets following the standard contrastive learning format. The positive sample is the original chunk used to generate a QA pair. For the AutoRAG and LlamaIndex datasets, the negative samples consist of two randomly selected chunks from the same corpus, which serve as 2 irrelevant context negatives. In contrast, for the RAGen dataset, two negative samples are used: one irrelevant context and one misleading context.

\paragraph{Results.}
Table~\ref{tab:retrieval_results} presents the complete results. All customized models outperform the un-customized baseline (denoted as Vanilla), confirming the necessity of domain-specific embedding customization. Datasets generated by RAGen consistently achieve superior performance across all domains and models, demonstrating the effectiveness of our data generation strategy.

\input{latex/my_table1}






\subsection{LLMs Supervised Fine-tuning}
\label{subsec:sft_lora}
\paragraph{Setup.} We perform standard LoRA-based supervised fine-tuning \cite{hu2022lora} on the Qwen2.5-1.5B and Qwen2.5-3B models \cite{qwen_qwen25_2025} using the QAC datasets generated from the three domains. All experiments are conducted using the open-source LlamaFactory framework \cite{zheng2024llamafactory}, with a fixed learning rate of 1e-5, five training epochs, and a 10\% validation split.

For input construction, we follow a consistent schema across all methods. In AutoRAG and LlamaIndex, the original chunk used to generate each question (the \textit{golden context}) is concatenated with the question to form the model input. For RAGen, all supportive evidence chunks are concatenated as the golden context.

To ensure a fair evaluation, we randomly sample 300 questions from the evaluation sets of each methods across all domains. Given that the task involves long-form QA, we adopt ROUGE-L and BERT-F1 as metrics for assessing lexical overlap and semantic similarity to evaluate model performance against reference answers.

\vspace{-6pt}
\input{latex/sft_lora}

\paragraph{Results.}Table~\ref{tab:finetune_results_15b3b} presents the results across all domains. Models fine-tuned on RAGen-generated data consistently outperform those trained on the AutoRAG and LlamaIndex datasets across both evaluation metrics—ROUGE-L and BERT-F1—thereby demonstrating superior factual consistency and semantic relevance.

These improvements validate the effectiveness of RAGen datasets. Notably, RAGen maintains its advantage across all three domains, indicating strong generalization ability beyond a single knowledge area. Furthermore, the consistent gains observed on both Qwen2.5-1.5B and Qwen2.5-3B confirm the scalability of our approach across model sizes.

\paragraph{Distractor Supervision Setup.}
Motivated by RAFT~\citep{RAG-LLM:RAFT}, which demonstrates the benefits of distractor exposure during training, we conduct additional experiments to evaluate how distractor-based supervision impacts LLM robustness in real-world RAG settings. We fine-tune models using both golden contexts and 2 distractors (irrelevant and misleading), and evaluate them using a fixed retriever with top-$k$=3 retrieved chunks on the customized embedding model trained in Sec.\ref{sec:embedding_adaptation}.

\paragraph{Results.}
Table~\ref{tab:qwen3b_apec_k3_nondist} presents the evaluation results on the PPFS domain using the Qwen-3B model and RAGen dataset. We observe a substantial performance drop when the model is fine-tuned without distractors and then exposed to noisy retrieved contexts during inference. In contrast, training with distractor-augmented supervision significantly improves robustness, yielding notable gains in both ROUGE-L and BERT-F1. These findings highlight the effectiveness of distractor-aware training in enhancing model resilience under realistic retrieval conditions.

\begin{table}[t!]
    \centering
    \footnotesize
    \begin{tabular}{lcc}
        \toprule
        \textbf{Method} & \textbf{ROUGE-L} & \textbf{BERT-F1} \\
        \midrule
        RAGen$_{\text{w/o dis}}$ & 0.3143 & 0.8957 \\
        RAGen$_{\text{dis}}$     & \textbf{0.4074} & \textbf{0.9121} \\
        \bottomrule
    \end{tabular}
    \vspace{0em}
    \caption{Evaluation of Qwen2.5-3B on the PPFS domain under real-world RAG inference ($k$=3) settings. 
    RAGen$_{\text{w/o dis}}$ is trained with golden contexts only, whereas 
    RAGen$_{\text{dis}}$ incorporates distractor supervision. All results are averaged over 3 runs.}
    \label{tab:qwen3b_apec_k3_nondist}
\end{table}

\section{Conclusion}
We present RAGen, a scalable and modular framework for generating high-quality, domain-specific QAC datasets to support diverse RAG adaptation strategies. Extensive experiments across multiple domains demonstrate its effectiveness in enhancing retrieval accuracy and answer quality, leading to more effective domain-adapted RAG systems. RAGen offers a practical solution for building domain-adapted RAG systems in complex, evolving knowledge environments.

\section*{Limitations}
While RAGen demonstrates strong performance in generating high-quality, domain-specific QAC datasets, several limitations remain.

First, the current pipeline operates only on text-formatted documents, whereas real-world enterprise knowledge often resides in PDFs or other multimodal formats (e.g., tables or images). Extending RAGen to robustly support multimodal inputs remains future work.

Second, the quality of seed documents directly affects the generated QAC samples; noisy or inconsistent sources may propagate errors into downstream adaptation.

Third, RAGen requires manual specification of the number of document-level concepts, a hyperparameter tied to document complexity. Automating this choice in a principled manner is an important direction for improvement.

Fourth, RAGen follows a bootstrapping pipeline in which errors in early retrieval or concept extraction may propagate to later stages. Incorporating more structured retrieval mechanisms, such as graph-based representations, could help mitigate this issue.

Finally, our evaluation relies on automatic and model-based metrics and does not include human studies, which may be necessary to fully assess qualities such as answer usefulness and faithfulness. We leave such evaluations to future work.
\bibliography{acl_latex}

\clearpage
\appendix
\section{Outline of the Appendix}
The appendix provides additional experimental analyses and implementation details that support the main results. It is organized as follows. Appendix \ref{app:qasper} presents an adaptation experiment on the QASPER dataset to evaluate the effectiveness of RAGen beyond enterprise-style corpora. Appendix \ref{app:k_sensitivity} analyzes the sensitivity of RAGen to the number of document-level concepts, and Appendix \ref{app:random_baseline} introduces a random multi-chunk baseline to isolate the impact of concept-guided evidence assembly. Finally, Appendix \ref{app:datasets} describes the dataset construction procedures, and Appendix \ref{app:prompt} reports the prompts used for concept extraction, question generation, and distractor construction.

\section{QASPER Adaptation Experiment}
\label{app:qasper}

To further examine whether RAGen is effective beyond custom enterprise-style corpora, we conduct an additional experiment on \textbf{QASPER} dataset \cite{Dasigi2021ADO}, a widely used benchmark for document-level question answering over scientific papers. This experiment serves two purposes: (i) to demonstrate that RAGen is applicable to general RAG datasets, and (ii) to verify that RAGen can still perform domain adaptation when the ``domain'' corresponds to a focused subfield rather than an organizational corpus.

We select 20 NLP-related papers from the QASPER dataset and treat them as a \emph{mini-domain} corpus. Following an adaptation-style evaluation protocol, we split the papers into 16 documents for adaptation (training) and 4 documents for evaluation.

From the 16 training documents, we generate synthetic QAC datasets using three methods: \textbf{RAGen} (our method), \textbf{AutoRAG}, and \textbf{LlamaIndex}. All methods use the same base LLM and embedding family, are executed on the same hardware and network, and generate comparable numbers of samples: 1030 (RAGen), 1,000 (AutoRAG), and 958 (LlamaIndex).

We then fine-tune the embedding model (BAAI/bge-large-en-v1.5) and the generator (Qwen2.5-3B-Instruct) on each synthetic dataset. Evaluation is performed on the ground-truth QASPER question--answer pairs associated with the selected papers.

\subsection{Retrieval Results}

Table~\ref{tab:qasper-retrieval} reports retrieval performance after fine-tuning the embedding model.

\begin{table}[H]
\centering
\small
\begin{tabular}{lcccc}
\toprule
Method & R@1 & R@5 & R@10 & MRR@10 \\
\midrule
Vanilla & 0.1452 & 0.3387 & 0.5323 & 0.2434 \\
AutoRAG & 0.1774 & 0.3871 & 0.5000 & 0.2687 \\
LlamaIndex & 0.1774 & 0.4032 & 0.5000 & 0.2639 \\
\textbf{RAGen} & \textbf{0.2581} & \textbf{0.4839} & \textbf{0.6129} & \textbf{0.3569} \\
\bottomrule
\end{tabular}
\caption{Retrieval performance on the QASPER NLP subset after embedding model fine-tuning on BAAI/bge-large-en-v1.5.}
\label{tab:qasper-retrieval}
\end{table}

RAGen yields substantial improvements across all metrics, particularly for R@1 and MRR@10, indicating more accurate and higher-ranked retrieval after adaptation.

\subsection{Generation Results}

Table~\ref{tab:qasper-generation} reports generation performance of the fine-tuned Qwen2.5-3B-Instruct model.

\begin{table}[H]
\centering
\small
\begin{tabular}{lcc}
\toprule
Method & ROUGE-L & BERT-F1 \\
\midrule
Vanilla & 0.2315 & 0.8524 \\
AutoRAG & 0.2423 & 0.8589 \\
LlamaIndex & 0.2401 & \textbf{0.8612} \\
\textbf{RAGen} & \textbf{0.2553} & \textbf{0.8612} \\
\bottomrule
\end{tabular}
\caption{Generation performance on the QASPER NLP subset after fine-tuning on Qwen2.5-3B-Instruct.}
\label{tab:qasper-generation}
\end{table}

RAGen achieves the best ROUGE-L score and matches the strongest baseline in BERT-F1, indicating improved answer overlap and semantic fidelity.

\subsection{Generation Time Analysis}

In addition to quality metrics, we report the wall-clock time required to generate synthetic QAC datasets for the QASPER NLP subset. All methods were executed on the same hardware and network environment.

Table~\ref{tab:qasper-runtime} reports the total time spent generating synthetic data for each method, together with the number of generated QACs.

\begin{table}[h]
\centering
\small
\begin{tabular}{lcc}
\toprule
\textbf{Method} & \textbf{\#QACs Generated} & \textbf{Time (minutes)} \\
\midrule
AutoRAG & 1,000 & 94 \\
LlamaIndex & 958 & 122 \\
RAGen & 1,030 & 134 \\
\bottomrule
\end{tabular}
\caption{Total generation time and number of generated QACs for each method on the QASPER NLP subset.}
\label{tab:qasper-runtime}
\end{table}

RAGen incurs additional offline cost due to concept extraction and concept-centered evidence assembly. In this experiment, approximately 93 minutes are spent constructing question stems (document-level concept extraction and sentence-level evidence selection), while the remaining 41 minutes are used for QAC generation once the stems are available. Importantly, this overhead is incurred only once per corpus and scales linearly with corpus size. Once constructed, question stems can be cached and reused for generating additional QACs at different difficulty levels or with alternative prompting strategies.

\section{Sensitivity Analysis on the Number of Document-Level Concepts}
\label{app:k_sensitivity}

In RAGen, the hyperparameter \(K\) controls the number of \emph{document-level concepts} extracted per document during the concept fusion stage (Sec.~3.1). Intuitively, larger values of \(K\) yield finer-grained semantic coverage and more potential question stems, while smaller values of \(K\) result in coarser representations with fewer training samples. We conduct a sensitivity analysis to examine the impact of \(K\) on downstream generation quality and to assess its suitability for large-scale adaptation.

\subsection{Experimental Setup}

We perform the analysis on the PPFS domain using \textit{Qwen2.5-3B-Instruct} as the generator. All experimental settings are kept identical except for the value of \(K\), which we vary over \(\{10, 15, 20, 25\}\). For each setting, we generate QAC datasets using RAGen and evaluate generation quality using ROUGE-L and BERT-F1.
Table~\ref{tab:k_sensitivity} reports the generation performance under different values of \(K\).

\begin{table}[H]
\centering
\small
\begin{tabular}{ccc}
\toprule
\(K\) & ROUGE-L & BERT-F1 \\
\midrule
10 & 0.3536 & 0.8907 \\
15 & 0.3815 & 0.9079 \\
20 & 0.3923 & 0.9039 \\
25 & 0.3905 & 0.9089 \\
\bottomrule
\end{tabular}
\caption{Generation performance on PPFS under different numbers of document-level concepts \(K\).}
\label{tab:k_sensitivity}
\end{table}

\subsection{Discussion}

Smaller values of \(K\) produce fewer document-level concepts and, consequently, fewer question stems and QACs, leading to slightly weaker generation quality. Increasing \(K\) initially improves performance by enabling richer semantic coverage and more diverse question construction. However, as \(K\) becomes large, performance gains diminish: because each document contains a finite amount of unique information, excessively large \(K\) values tend to yield overlapping or weakly informative concepts, providing limited additional benefit.

Overall, performance is relatively stable for \(K \in [15, 25]\), indicating that RAGen is not overly sensitive to this hyperparameter. We adopt \(K = 15\) as a default setting in our main experiments, as it provides a favorable trade-off between semantic coverage, dataset size, and computational cost, making it well-suited for large-scale domain adaptation.

\section{Random Multi-Chunk Baseline}
\label{app:random_baseline}

To examine whether the performance gains of RAGen arise solely from the use of multiple chunks, rather than from its concept-guided evidence assembly, we introduce an additional \emph{Random multi-chunk} baseline. This baseline provides a stronger comparison point than standard single-chunk question generation pipelines.

\subsection{Baseline Construction}

For the Random multi-chunk baseline, we randomly sample two \emph{non-adjacent} chunks from the same document and concatenate them to form a multi-chunk context. Question--answer pairs are then generated from this concatenated context using the standard AutoRAG pipeline. This design ensures that the baseline has access to multi-chunk information, while avoiding any semantic clustering or concept-based guidance in evidence selection.

We generate synthetic QAC datasets using this Random baseline on the \textbf{PPFS} domain, and fine-tune both the retriever and generator using the same protocol as in our main experiments. Specifically, we fine-tune \textit{BAAI/bge-large-en-v1.5} for retrieval and \textit{Qwen2.5-3B-Instruct} for generation.

\subsection{Retrieval and Generation Results}

Table~\ref{tab:random-retrieval} reports retrieval performance after fine-tuning the embedding model.

\begin{table}[h]
\centering
\small
\begin{tabular}{lcccc}
\toprule
Method & R@1 & R@5 & R@10 & MRR@10 \\
\midrule
Vanilla & 0.1548 & 0.4348 & 0.5549 & 0.2722 \\
AutoRAG & 0.1877 & 0.5183 & 0.6712 & 0.2247 \\
LlamaIndex & 0.2024 & 0.5604 & 0.6987 & 0.3548 \\
Random & 0.1895 & 0.5535 & 0.6816 & 0.2337 \\
\textbf{RAGen} & \textbf{0.3095} & \textbf{0.6584} & \textbf{0.7821} & \textbf{0.4626} \\
\bottomrule
\end{tabular}
\caption{Retrieval performance on PPFS after retriever fine-tuning using different synthetic datasets.}
\label{tab:random-retrieval}
\end{table}

The Random baseline improves over single-chunk generation in some metrics, confirming that access to multi-chunk context can be beneficial. However, RAGen substantially outperforms Random across all retrieval metrics, particularly for R@1 and MRR@10.

Table~\ref{tab:random-generation} reports generation performance of the fine-tuned \textit{Qwen2.5-3B-Instruct} model.

\begin{table}[h]
\centering
\small
\begin{tabular}{lcc}
\toprule
Method & ROUGE-L & BERT-F1 \\
\midrule
AutoRAG & 0.3436 & 0.8979 \\
LlamaIndex & 0.3253 & 0.8952 \\
Random & 0.3634 & 0.8962 \\
\textbf{RAGen} & \textbf{0.3815} & \textbf{0.9079} \\
\bottomrule
\end{tabular}
\caption{Generation performance on PPFS after generator fine-tuning.}
\label{tab:random-generation}
\end{table}

\subsection{Discussion on random baseline}

The Random multi-chunk baseline shows that naive concatenation of multiple chunks can improve performance relative to single-chunk generation. However, its gains are limited and inconsistent, as random concatenation often introduces long and noisy contexts containing irrelevant information.

In contrast, RAGen consistently outperforms Random by constructing \emph{concept-centered} and \emph{sentence-level} evidence sets that preserve cross-chunk reasoning while maintaining focused and compact contexts. These results indicate that the improvements of RAGen primarily stem from concept-guided evidence assembly, rather than from multi-chunk input alone.

\section{Dataset Construction}
\label{app:datasets}

We construct three domain-specific corpora from publicly available sources, each representing a realistic deployment scenario for domain-adapted RAG systems. All documents are collected from official or well-established public websites.

\paragraph{PPFS.}
The PPFS corpus is derived from publications of the \emph{APEC Policy Partnership on Food Security (PPFS)}. We collected 18 policy and meeting documents covering topics such as food security, water management, rural development, and sustainable agriculture. Documents were obtained from the official APEC publications portal:\footnote{\url{https://www.apec.org/publications}}.

\paragraph{TradePolicy.}
The TradePolicy corpus consists of import and export regulations (primarily for meat and seafood products) collected from official government portals of eight APEC economies. Example sources include the Singapore Food Agency\footnote{\url{https://www.sfa.gov.sg/food-import-export}} and the Thailand FDA\footnote{\url{https://en.fda.moph.go.th/entrepreneurs-food}}.

\paragraph{BusinessAI.}
The BusinessAI corpus comprises technical and analytical reports on the adoption of artificial intelligence across business sectors. We collected 20 public articles from the McKinsey website by querying the official website with the keyword \emph{“Artificial Intelligence”}\footnote{\url{https://www.mckinsey.com/search}}. Selected documents focus on real-world case studies and organizational deployments of AI technologies.

We will release the processed versions of these dataset, and  plan to expand these corpora in future work to support larger-scale and longitudinal studies of domain-adapted RAG systems.

\section{Prompts}
\label{app:prompt}
\begin{figure}[htbp]
\centering
\begin{tcolorbox}[width=\columnwidth, colback=gray!6, colframe=black,
                  boxsep=4pt, left=4pt, right=4pt, top=4pt, bottom=4pt]
\small
\textbf{Instructions:} You are an expert analyst specializing in distilling complex documents into structured concept maps.  
Given the following excerpt from a longer document, extract the main concepts, and supporting details that are critical to understanding the material.\\
\textbf{Excerpt:} <start>{excerpt}<end>
\end{tcolorbox}
\caption{Chunk level concept extraction prompt.}
\end{figure}

\begin{figure}[htbp]
\centering
\begin{tcolorbox}[width=\columnwidth, colback=gray!6, colframe=black,
                  boxsep=4pt, left=4pt, right=4pt, top=4pt, bottom=4pt]
\small
\textbf{Instructions:} You are an expert question generation model for a domain-specific Retrieval-Augmented Generation (RAG) pipeline.
Your task is to generate diverse, high-quality question–answer pairs based ONLY on a given TOPIC and its associated EVIDENCES.

Every QA pair must be fully supported by the evidence provided.
If the evidence is insufficient to form any valid question–answer pairs, return an empty list. \\
\textbf{Workflow}:
1. Use Bloom’s Revised Taxonomy as overall guidance when generating questions:
   \\- C1 Remember: Recall or recognize facts.
   \\- C2 Understand: Explain or interpret meaning.
   \\- C3 Apply: Use information in a scenario.
   \\- C4 Analyze: Compare, contrast, or relate parts.
   \\- C5 Evaluate: Make judgments with reasons.
   \\- C6 Create: Combine or reorganize ideas into new forms.

2. Generate a diverse set of realistic, natural-sounding questions that can be asked based on the provided evidences.
    - Cover Bloom’s levels (C1–C6) as much as possible.
    - Prioritize deeper reasoning and higher-order questions (C4–C6) when the evidence supports them.

3. After question generated, assign the most appropriate Bloom’s cognitive level (C1–C6) based on the complexity of the required reasoning.
\end{tcolorbox}
\caption{Bloom guided single-stem question generation prompt.}
\end{figure}
\label{sec:single_stem_prompt}

\begin{figure}[htbp]
\centering
\begin{tcolorbox}[width=\columnwidth, colback=gray!6, colframe=black,
                  boxsep=4pt, left=4pt, right=4pt, top=4pt, bottom=4pt]
\small
\textbf{Instructions:} You are an expert question generation model for a domain-specific Retrieval-Augmented Generation (RAG) pipeline.
Your task is to generate diverse, high-quality question–answer pairs based ONLY on a set of given TOPICs and their associated EVIDENCES.
Every QA pair must be fully supported by the evidence provided.
All questions and answers must be strictly grounded in the evidence provided — no assumptions or world knowledge are allowed.
If the evidence is insufficient or the topics are semantically unrelated, return an empty list: []

\textbf{Workflow:}
1. Use Bloom’s Revised Taxonomy as overall guidance when generating questions:
    \\ - C1 Remember: Recall or recognize facts.
    \\ - C2 Understand: Explain or interpret meaning.
    \\ - C3 Apply: Use information in a scenario.
    \\ - C4 Analyze: Compare, contrast, or relate parts.
    \\ - C5 Evaluate: Make judgments with reasons.
    \\ - C6 Create: Combine or reorganize ideas into new forms.
2. Identify meaningful conceptual relations among the given topics before forming any question.
    (a) Compute whether there is a shared entity, process, cause–effect link, or co-occurring theme between topics and their associated evidences.
    (b) If no such overlap exists, immediately return [].

\end{tcolorbox}
\caption{Multi-stem question generation prompt.}
\end{figure}
\label{sec:single_stem_prompt}

\begin{figure}[H]
\centering
\begin{tcolorbox}[width=\columnwidth, colback=gray!6, colframe=black,
                  boxsep=4pt, left=4pt, right=4pt, top=4pt, bottom=4pt]
\small
\textbf{Instructions:} Given a question, your task is to generate one distractor context that:
	1.	Looks topically related to the question,
	2.	Does NOT help answer the question,
	3.	Could mislead someone who reads quickly or carelessly,
	4.	Avoids including the correct answer or direct hints to it.
\end{tcolorbox}
\caption{Dis-tractor context generation prompt.}
\end{figure}
\label{sec:appendix}

\end{document}

%% file: latex/my_table1.tex
\begin{table*}[t!]
    \centering
    \resizebox{\textwidth}{!}{
    \begin{tabular}{ll cccc cccc cccc}
        \toprule
        \multirow{2}{*}{\textbf{Vanilla Model}} & \multirow{2}{*}{\textbf{Finetune Strategy}} & \multicolumn{4}{c}{\textbf{PPFS}} & \multicolumn{4}{c}{\textbf{TradePolicy}} & \multicolumn{4}{c}{\textbf{BusinessAI}} \\
        \cmidrule(lr){3-6} \cmidrule(lr){7-10} \cmidrule(lr){11-14}
         &  & R@1 & R@5 & R@10 & MRR@10 & R@1 & R@5 & R@10 & MRR@10 & R@1 & R@5 & R@10 & MRR@10 \\
        \midrule
        \multirow{5}{*}{BGE-large\cite{bge-large-v15}}
         & Vanilla              & 0.1548 & 0.4368 & 0.5549 & 0.2722 & 0.1961 & 0.4691 & 0.6214 & 0.3154 & 0.1068 & 0.3291 & 0.4263 & 0.2019 \\
         & AutoRAG         & 0.1877 & 0.5183 & 0.6712 & 0.3342 & 0.2247 & 0.5505 & 0.6606 & 0.3573 & 0.1560 & 0.4818 & 0.6325 & 0.2972 \\
         & LlamaIndex      & 0.2024 & 0.5604 & 0.6987 & 0.3548 & 0.2474 & 0.5686 & 0.6893 & 0.3789 & 0.1624 & 0.4893 & 0.6261 & 0.3036 \\
         & RAGen       & \textbf{0.3095} & \textbf{0.6584} & \textbf{0.7821} & \textbf{0.4626} & \textbf{0.3891} & \textbf{0.8069} & \textbf{0.8899} & \textbf{0.5586} & \textbf{0.3002} & \textbf{0.6827} & \textbf{0.8120} & \textbf{0.4693} \\
        \midrule
        \multirow{5}{*}{BGE-m3\cite{chen2024bge}}
         & Vanilla              & 0.2115 & 0.5018 & 0.6136 & 0.3359 & 0.2368 & 0.5309 & 0.6516 & 0.3584 & 0.1368 & 0.4241 & 0.5417 & 0.2602 \\
         & AutoRAG         & 0.2015 & 0.5055 & 0.6383 & 0.3377 & 0.2594 & 0.5807 & 0.6953 & 0.3909 & 0.1603 & 0.5043 & 0.6271 & 0.3066 \\
         & LlamaIndex      & 0.2125 & 0.5687 & 0.7042 & 0.3664 & 0.2881 & 0.5792 & 0.7074 & 0.4114 & 0.1538 & 0.4947 & 0.6282 & 0.3000 \\
         & RAGen       & \textbf{0.2692} & \textbf{0.6255} & \textbf{0.7647} & \textbf{0.4261} & \textbf{0.3665} & \textbf{0.7888} & \textbf{0.8944} & \textbf{0.5355} & \textbf{0.2318} & \textbf{0.6677} & \textbf{0.7906} & \textbf{0.4232} \\
        \midrule
        \multirow{5}{*}{E5-large-v2\cite{wang2022text}}
         & Vanilla              & 0.1749 & 0.4844 & 0.6273 & 0.3052 & 0.1131 & 0.4449 & 0.5913 & 0.2472 & 0.1015 & 0.3205 & 0.4573 & 0.1977 \\
         & AutoRAG         & 0.1905 & 0.5201 & 0.6465 & 0.3274 & 0.1388 & 0.4449 & 0.6199 & 0.2685 & 0.1047 & 0.3226 & 0.4679 & 0.2049 \\
         & LlamaIndex      & 0.1996 & 0.5348 & 0.6767 & 0.3451 & 0.1976 & 0.5158 & 0.6440 & 0.3259 & 0.1026 & 0.3568 & 0.4979 & 0.2123 \\
         & RAGen       & \textbf{0.2665} & \textbf{0.6511} & \textbf{0.7848} & \textbf{0.4345} & \textbf{0.3469} & \textbf{0.7677} & \textbf{0.8778} & \textbf{0.5074} & \textbf{0.2767} & \textbf{0.6912} & \textbf{0.8066} & \textbf{0.4554} \\
        \bottomrule
    \end{tabular} 
    }
    \caption{Retrieval performance on 3 domains. the best results are in bold. All results are averaged over 3 runs.}
    \vspace{-4pt}
    \label{tab:retrieval_results}
    \vspace{-8pt}
\end{table*}

%% file: latex/sft_lora.tex
\begin{table}[b!]
    \centering
    \footnotesize
    \renewcommand{\arraystretch}{0.85}
    \resizebox{0.47\textwidth}{!}{
    \begin{tabular}{llcc}
        \toprule
        \textbf{Domain} & \textbf{Method} & \textbf{ROUGE-L} & \textbf{BERT-F1} \\
        \midrule
        \multicolumn{4}{c}{\textbf{Qwen2.5–1.5B Instruct}} \\
        \midrule
        \multirow{3}{*}{PPFS}
          & AutoRAG      & 0.2876 & 0.8847 \\
          & LlamaIndex   & 0.3293 & 0.8903 \\
          & RAGen        & \textbf{0.3955} & \textbf{0.9094} \\
        \cmidrule(lr){1-4}
        \multirow{3}{*}{TradePolicy}
          & AutoRAG      & 0.2775 & 0.8726 \\
          & LlamaIndex   & 0.2698 & 0.8696 \\
          & RAGen        & \textbf{0.3911} & \textbf{0.9033} \\
        \cmidrule(lr){1-4}
        \multirow{3}{*}{BusinessAI}
          & AutoRAG      & 0.2701 & 0.8852 \\
          & LlamaIndex   & 0.3223 & 0.8925 \\
          & RAGen        & \textbf{0.3392} & \textbf{0.9038} \\
        \midrule
        \multicolumn{4}{c}{\textbf{Qwen2.5–3B Instruct}} \\
        \midrule
        \multirow{3}{*}{PPFS}
          & AutoRAG      & 0.3436 & 0.8979 \\
          & LlamaIndex   & 0.3253 & 0.8952 \\
          & RAGen        & \textbf{0.3815} & \textbf{0.9079} \\
        \cmidrule(lr){1-4}
        \multirow{3}{*}{TradePolicy}
          & AutoRAG      & 0.3388 & 0.8875 \\
          & LlamaIndex   & 0.3346 & 0.8861 \\
          & RAGen        & \textbf{0.3747} & \textbf{0.9004} \\
        \cmidrule(lr){1-4}
        \multirow{3}{*}{BusinessAI}
          & AutoRAG      & 0.3284 & 0.8985 \\
          & LlamaIndex   & 0.3597 & 0.9036 \\
          & RAGen        & \textbf{0.3682} & \textbf{0.9091} \\
        \bottomrule
    \end{tabular}
    }
    \caption{Performance comparison of Qwen2.5–1.5B and –3B models on 3 domains. All results are averaged over 3 runs. The best result is in bold.}
    \label{tab:finetune_results_15b3b}
\end{table}